\documentclass[runningheads]{llncs}
\usepackage{graphicx}
\usepackage{listings}
\usepackage{amsmath,amsfonts,amssymb}
\usepackage{bm}
\usepackage{enumitem}
\usepackage{booktabs,makecell}
\usepackage{multirow}

\newcommand{\twolinecell}[2][c]{%
  \begin{tabular}[#1]{@{}c@{}}#2\end{tabular}}

\begin{document}
\title{Interpreting and Correcting Medical Image Classification with PIP-Net}
\author{Meike Nauta\inst{1}\orcidID{0000-0002-0558-3810} \and
Johannes H. Hegeman\inst{1,2}\orcidID{0000-0003-2188-2738} \and Jeroen Geerdink\inst{2}\orcidID{0000-0001-6718-6653} \and Jörg Schlötterer\inst{3,4}\orcidID{0000-0002-3678-0390} \and Maurice van Keulen\inst{1} \and
Christin Seifert\inst{1,4}\orcidID{0000-0002-6776-3868}}
\authorrunning{M. Nauta et al.}
\institute{University of Twente, Enschede, the Netherlands \and
Hospital Group Twente, Almelo \& Hengelo, the Netherlands \and
University of Mannheim, Germany \and University of Marburg, Hessian.AI, Marburg, Germany\\
\email{\{m.nauta,m.vankeulen\}@utwente.nl, \{h.hegeman,j.geerdink\}@zgt.nl, \{joerg.schloetterer,christin.seifert\}@uni-marburg.de}}
\maketitle              % typeset the header of the contribution
\begin{abstract}
Part-prototype models are explainable-by-design image classifiers, and a promising alternative to black box AI. This paper explores the applicability and potential of interpretable machine learning, in particular PIP-Net, for automated diagnosis support on real-world medical imaging data. PIP-Net learns human-understandable prototypical image parts and we evaluate its accuracy and interpretability for fracture detection and skin cancer diagnosis. 
We find that PIP-Net's decision making process is in line with medical classification standards, while only provided with image-level class labels. Because of PIP-Net's unsupervised pretraining of prototypes, data quality problems such as undesired text in an X-ray or labelling errors can be easily identified. Additionally, we are the first to show that humans can manually correct the reasoning of PIP-Net by directly disabling undesired prototypes. We conclude that part-prototype models are promising for medical applications due to their interpretability and potential for advanced model debugging.

\end{abstract}
\setcounter{footnote}{0}
\section{Introduction}
Deep learning has shown great promise in medical imaging tasks, as neural networks can outperform clinicians in fracture detection~\cite{langerhuizen_what_2019} or have equivalent performance in medical diagnosis~\cite{liu_comparison_2019}. 
Machine learning (ML) models are usually evaluated in terms of predictive performance, e.g., classification accuracy. However, performance metrics do not capture whether the evaluated model is right for the right reasons~\cite{lapuschkin_unmasking_2019}. ML models can replicate biases and other confounding patterns from the input data when these are discriminative for the downstream task. For example, COVID-19 detectors were found to rely on markers, image edges, arrows and other annotations in chest X-rays~\cite{degrave2021ai_covid_shortcuts}, and 
\cite{badgeley_deep_2019} showed that deep learning models can predict hip fractures indirectly through confounding patient and healthcare variables rather than directly detecting the fracture in the image. 
This so-called ``Clever Hans'' behaviour~\cite{lapuschkin_unmasking_2019} makes a medical classifier right for the \emph{wrong} reasons, by basing its predictions on clinically irrelevant artefacts that are not representative for the actual data distribution~\cite{geirhos2020shortcut,ijcai/RossHD17}. Such shortcut learning will therefore lead to a lack of generalisation and in turn unsatisfactory performance once deployed in clinical practice~\cite{geirhos2020shortcut}. 

Detecting shortcut learning and other undesired reasoning is challenging since neural networks are black boxes~\cite{adadi_peeking_2018}. 
Explainable AI (XAI) aims to provide insight into  the reasoning of a predictive model.
A commonly used explanation method in medical imaging is the feature attribution map~\cite{borys_nauta_xai_saliency_2023}, a heatmap that highlights relevant regions in an input image~\cite{Salahuddin_2022_transparency}. These explanation methods are, however, \emph{post-hoc}, i.e., they reverse-engineer an already trained predictive model. With the explanation method detached from the predictive model, the explanation is not guaranteed to truthfully mimic the internal calculations of the black box. Additionally, such heatmaps do not explain the \emph{full} reasoning process but only give an intuition, making them irrelevant to tasks in realistic scenario's~\cite{rudin2019stop,Shen_Huang_2020_howuseful,colin_what_2022}. Specifically, it has been shown that feature attribution maps do not fulfil clinical requirements to correctly explain a model's decision process~\cite{Jin_Li_Hamarneh_2022}. 

Recently, as an alternative to post-hoc explanations, intrinsically interpretable models are proposed based on prototypical parts~\cite{nauta2020neural,nips/ChenLTBRS19,nauta_pipnet,rymarczyk_2022_protopool}. Their reasoning follows the recognition-by-components theory~\cite{biederman1987recognition} by analysing whether patches in an input image are similar to a learned prototypical part. Importantly, part-prototype models do not require any part annotations and only rely on image-level class labels. Most of these models have been developed for fine-grained natural image recognition, including recognising bird species and car types. Only a few works apply part-prototype models to medical images: ProtoPNet~\cite{nips/ChenLTBRS19} is also applied to chest X-rays~\cite{singh_2021_interpretable} and MRI scans~\cite{mohammadjafari2021using}, ProtoMIL~\cite{rymarczyk2021protomil} is developed for histology slide classification and \cite{barnett2021case} adapted ProtoPNet for mammography by including a loss based on fine-grained expert image annotations. 

In this work, we show the applicability and potential of PIP-Net for understanding and \emph{correcting} medical imaging classification, and thus contribute towards explanatory interactive model debiasing~\cite{teso_2019_explantory_interactive,ANDERS2022261,pahde2023reveal}. PIP-Net~\cite{nauta_pipnet} is a next-generation interpretable ML method that lets users understand its sparse reasoning with prototypical parts and will abstain from a decision for out-of-distribution input. In addition, its intuitive design empowers users to make adjustments in the model's reasoning. We investigate to what extent PIP-Net can be used for revealing and correcting \emph{shortcut learning} when applied to fracture detection and skin cancer diagnosis. Generally, one can identify three ways of debugging models: either by adapting the dataset to neutralise the bias (e.g.~\cite{nauta_2022_skinshortcut}), by adapting the model's loss function~(e.g.~\cite{barnett2021case,icml/RiegerSMY20}), or by adapting the predictive model directly. Steering a model's reasoning through a loss function requires manual annotations that indicate where a model should or should not focus. Although such additional annotations could improve predictive accuracy, providing such annotations by physicians is complex and time-consuming, and therefore expensive and often unfeasible. Instead, the interpretable scoring sheet of PIP-Net allows debugging the model directly by simply disabling shortcut prototypes. 
% Hence, PIP-Net contributes towards `explanatory interactive ML' which can improve predictive and explanatory powers~\cite{teso_2019_explantory_interactive} and allows bias corrections~\cite{ANDERS2022261,pahde2023reveal}.

In summary, we show that PIP-Net applied to medical imaging:
\begin{itemize}[nosep]
\item learns an interpretable, sparse scoring sheet with semantically meaningful prototypical parts~(Sections~\ref{ssec_medpipnet_performance_sparsity} and~\ref{ssec:medpipnet_perceived_purity}),
\item can reveal data quality problems, including spurious artefacts and labelling errors~(Section~\ref{ssec:medpipnet_dat_quality}),
    \item learns interpretable reasoning that is in line with medical domain knowledge and classification standards~(Section~\ref{sec:medpipnet_alignment_med}),
    \item can reveal shortcut learning and subsequently be `debugged' by clinicians by disabling undesired prototypes~(Section~\ref{sec:medpipnet_reveal_correct_shortcut_learning}).
\end{itemize}

\section{Background on PIP-Net}
\begin{figure*}
    \centering
    \includegraphics[width=\linewidth]{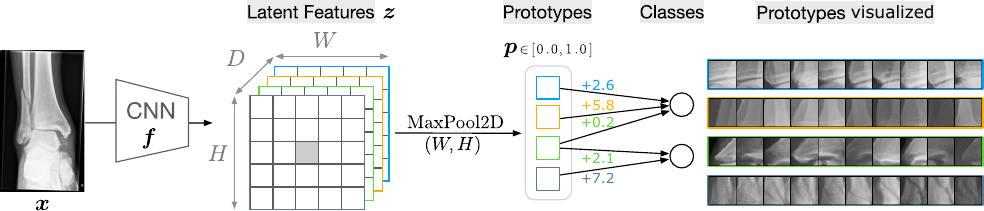}
    \caption{PIP-Net consists of a CNN backbone (e.g. ConvNeXt with $D=768$) with a novel training mechanism to disentangle the latent space and learn prototypical representations $\bm{z}$. The feature representations are pooled to a vector of prototype presence scores $\bm{p}$. Prototypes and classes are connected via a sparse linear layer. Model outputs during inference are not normalised and allow the outputs to be interpreted as a simple scoring sheet. A prototype can be visualised as an image patch by upsampling the corresponding patch in $\bm{z}$.}
    \label{fig:pipnet_med_overview}
\end{figure*}

PIP-Net~\cite{nauta_pipnet} is a deep learning model designed with interpretability integrated into its architecture and training mechanism. It consists of a convolutional neural network (CNN) with loss terms that disentangle the latent space and optimise the model to learn semantically meaningful components, while only having access to image-level class labels and thus not relying on additional part annotations. The learned components are ``prototypical parts'' (prototypes) which are visualised as image patches. Subsequently, PIP-Net classifies images by connecting the learned prototypes to classes via a sparse, linear layer. The linear decision making process is therefore globally interpretable as a scoring sheet where the presence of a prototypical part in an input image adds to the evidence for a particular class. In case no relevant part-prototypes are found, all output scores stay zero. Hence, PIP-Net can abstain from a decision for out-of-distribution input. 
An overview of the architecture is shown in Figure~\ref{fig:pipnet_med_overview}. In contrast to ProtoPNet~\cite{nips/ChenLTBRS19} that requires a fixed number of prototypes per class, PIP-Net has a novel training mechanism and is optimised for sparsity \emph{and} accuracy. It therefore learns the suitable number of prototypes itself. Training of PIP-Net consists of two stages: in the first stage, the latent space is automatically disentangled into prototypes that are learned self-supervised with contrastive learning. In this stage, no image labels are needed and hence would allow the use of additional unlabelled data, which is highly relevant due to expensive labelling effort in the medical domain. The second training stage learns the weights in the sparse classification layer while finetuning the prototypes based on image-level labels. 

\section{Datasets and Experimental Setup}
We evaluate the predictive accuracy and interpretability of 
PIP-Net with two open benchmark datasets on skin cancer diagnosis (ISIC~\cite{codella_skin_2019}) and bone abnormality detection in bone X-rays (MURA~\cite{rajpurkar2017mura}), and two real-world data sets from a Dutch hospital on hip and ankle fracture detection (HIP resp. ANKLE). Table~\ref{tab:data_stats} shows descriptions of the datasets and Fig.~\ref{fig:data_imgs_examples} some example images. 

\paragraph{Data Preprocessing:}
We sample a random split of 20\% of the HIP and ANKLE datasets as test sets. 
For ISIC, we use the same 361 malignant images in the test set as~\cite{nauta_2022_skinshortcut} and randomly sample a similar fraction from the benign images. The MURA dataset contains a fixed training and validation split (which we use as test set as the hidden test set competition is closed). All images are resized to $224 \times 224$ and augmented with TrivialAugment~\cite{Muller_2021_ICCV_trivialaugment}. We account for imbalanced data with class over-sampling using a weighted sampler. Code including the data preprocessing is available at \url{https://github.com/M-Nauta/PIPNet}.

In radiography, usually multiple X-rays (with different views) are taken for one \emph{study}. Therefore, images in MURA, HIP and ANKLE are annotated at the study level. In this work, we assign the study label to each image individually and prevent data leakage by splitting data in train and test sets \emph{per study}. Conceptually, we think that PIP-Net could also be used for interpretable multi-instance classification by doing a max-pooling operation over all latent image representations $\bm{z}$ in a study, rather than a single image embedding~(see Fig.~\ref{fig:pipnet_med_overview}). PIP-Net can then indicate in which image and at which location evidence for a certain class is found. We leave experimental investigation on multi-instance PIP-Net for future work. 

\begin{table*}[t!h]
    \centering
    \addtolength{\tabcolsep}{-1pt}
    % \small{
    \caption{Description of the datasets, including their number of samples. All used datasets are binary labelled (indicating  presence or absence of abnormality).}
\label{tab:data_stats}
    \begin{tabular}{p{0.02\linewidth}@{\hskip 6pt}p{0.75\linewidth}@{\hskip 6pt}>{\centering\arraybackslash}p{0.22\linewidth}}
    \toprule
         & \textbf{Description} & \twolinecell[t]{\textbf{Label} \\ present / absent} \\
        \midrule
         \multirow{5}{*}{\bf \rotatebox{90}{ISIC}} & \small{Public skin cancer dataset from the International Skin Imaging Collaboration (ISIC)~\cite{codella_skin_2019}. Specifically selected as it is known to contain confounding artefacts~\cite{icml/RiegerSMY20,nauta_2022_skinshortcut}: half of the images of benign lesions contain elliptical, coloured patches (colour calibration charts~\cite{mishra2016overview}), whereas the malignant lesion images contain none.} & \small{\textbf{\# Train images:} \newline 2,192 / 16,998 \newline \textbf{\# Test images:} \newline 361 / 2,736}\\\midrule
         \multirow{6}{*}{\bf \rotatebox{90}{MURA}} &  \small{Public dataset with musculoskeletal radiographs~\cite{rajpurkar2017mura} from different body parts, including the shoulder, humerus, elbow, forearm, wrist, hand, and finger. The bone X-rays are labelled as `normal' or `abnormal'. Images with the `abnormal' label have any abnormal finding, including the presence of a fracture, hardware, lesions and joint diseases.} & \small{\textbf{\# Train images:} \newline 14,873 / 21,935 \newline \textbf{\# Test images:}  1,530 / 1,667}\\\midrule
          \multirow{9}{*}{\bf \rotatebox{90}{HIP}} &  \small{Dataset from hospital Ziekenhuisgroep Twente (ZGT). Included were hip/pelvis radiographic studies (2005-2018, patients $\geq$ 21y old). Studies were labelled based on an administrative code and by analysing radiology reports with a rule-based approach. Images from follow-up studies were excluded. A sample of the selected studies (127 with fracture and 204 without) was manually verified by a radiologist. Images were anonymised by removing Protected Health Information (PHI). All radiographic images were converted from DICOM format~\cite{kahn2007dicom} to JPG.} & \small{\textbf{\# Train images:} \newline 3,468 / 4,080  \newline \textbf{\# Test images:}  859 / 1,005}\\\midrule
          \multirow{4}{*}{\bf \rotatebox{90}{ANKLE}} & \small{Dataset from ZGT. Selected were ankle studies (2005-2020) based on administrative code for ``ankle fracture'' or ``ankle distortion'' (no fracture). Images from follow-up studies were excluded, PHI information removed and images converted from DICOM to JPG.}  & \small{\textbf{\# Train images:} \newline 12,233 / 8,602 \newline \textbf{\# Test images:}  3,033 / 2,169}\\
         \bottomrule
    \end{tabular} 
\end{table*} 
\normalsize

\paragraph{Model Training:}
We use a ConvNeXt-tiny~\cite{convnext_Liu_2022_CVPR} backbone, adapted to output feature maps of size $13 \times 13$ and pretrained on ImageNet, as it is shown that CNNs pretrained on natural images and with adequate fine-tuning outperform, or perform just as well as, CNNs trained for radiology from scratch~\cite{tajbakhsh_convolutional_2016}. 
Similar to the training process for PIP-Net on natural images, we train PIP-Net with a learning rate of 0.05 for the linear classification layer, and 0.0001 for the backbone. 
We use a batch size of 64 and adapt the number of epochs, such that the number of weight updates is similar for all datasets. We apply roughly 2,000 updates for pretraining the prototypes and 10,000 updates for the second training stage.
This calculation results in 6 (pretrain) and 34 (training) epochs for ISIC, 4 and 18 epochs for MURA, 16 and 85 epochs for HIPS and 6 and 31 epochs for ANKLE. Results reported in this paper are based on a slightly older version of PIP-Net. On natural images, the final published version of PIP-Net gives similar or higher prediction accuracy and similar prototypes when compared to this older version, hence we don't expect significant differences.

\begin{figure*}[th]
    \centering
\setlength\tabcolsep{1.0pt} 
    \begin{tabular}{cp{0.12\linewidth}p{0.12\linewidth}p{0.12\linewidth}p{0.12\linewidth}p{0.004\linewidth}p{0.12\linewidth}p{0.12\linewidth}p{0.12\linewidth}p{0.12\linewidth}}
         & \multicolumn{4}{c}{Malignant} & &\multicolumn{4}{c}{Benign}  \\
         \rotatebox[origin=c]{90}{\bf ISIC} & \parbox[c]{\linewidth}{\includegraphics[width=\linewidth]{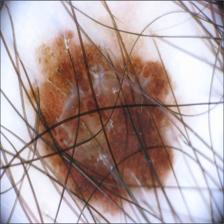}} & \parbox[c]{\linewidth}{\includegraphics[width=\linewidth]{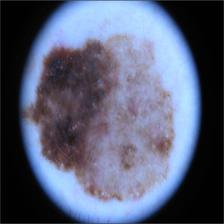}} & \parbox[c]{\linewidth}{\includegraphics[width=\linewidth]{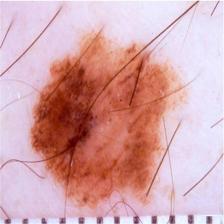}}&\parbox[c]{\linewidth}{\includegraphics[width=\linewidth]{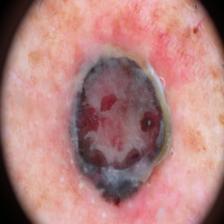}}&  &
          \parbox[c]{\linewidth}{\includegraphics[width=\linewidth]{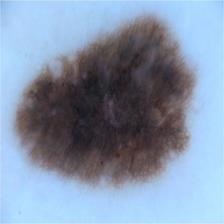}} & \parbox[c]{\linewidth}{\includegraphics[width=\linewidth]{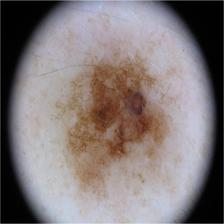}} & \parbox[c]{\linewidth}{\includegraphics[width=\linewidth]{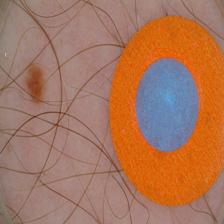}}&\parbox[c]{\linewidth}{\includegraphics[width=\linewidth]{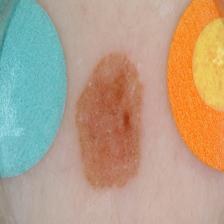}}\\
         
         & \multicolumn{4}{c}{Abnormal} & &\multicolumn{4}{c}{Normal}  \\
         \rotatebox[origin=c]{90}{\bf MURA} & \parbox[c]{\linewidth}{\includegraphics[width=\linewidth]{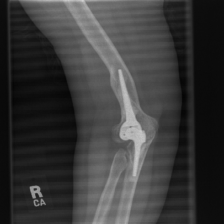}} & \parbox[c]{\linewidth}{\includegraphics[width=\linewidth]{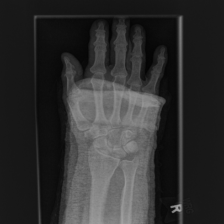}} & \parbox[c]{\linewidth}{\includegraphics[width=\linewidth]{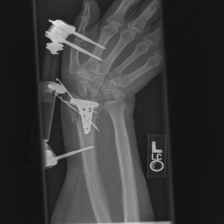}}&\parbox[c]{\linewidth}{\includegraphics[width=\linewidth]{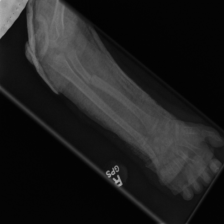}}&  &
          \parbox[c]{\linewidth}{\includegraphics[width=\linewidth]{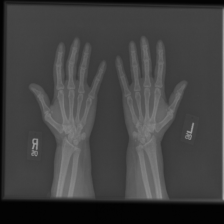}} & \parbox[c]{\linewidth}{\includegraphics[width=\linewidth]{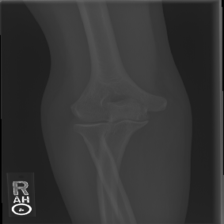}} & \parbox[c]{\linewidth}{\includegraphics[width=\linewidth]{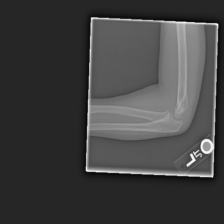}}&\parbox[c]{\linewidth}{\includegraphics[width=\linewidth]{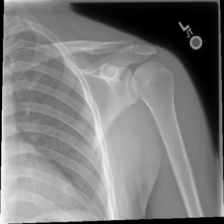}}\\
         & \multicolumn{4}{c}{Fracture} & &\multicolumn{4}{c}{No Fracture} \\
         \rotatebox[origin=c]{90}{\bf HIP} & \parbox[c]{\linewidth}{\includegraphics[width=\linewidth]{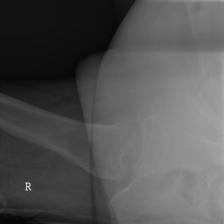}} & \parbox[c]{\linewidth}{\includegraphics[width=\linewidth]{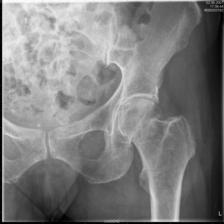}} & \parbox[c]{\linewidth}{\includegraphics[width=\linewidth]{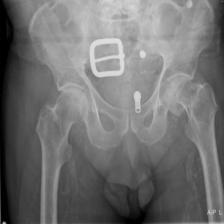}}&\parbox[c]{\linewidth}{\includegraphics[width=\linewidth]{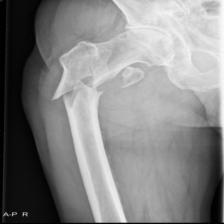}} &  & \parbox[c]{\linewidth}{\includegraphics[width=\linewidth]{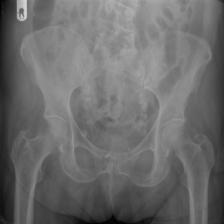}} & \parbox[c]{\linewidth}{\includegraphics[width=\linewidth]{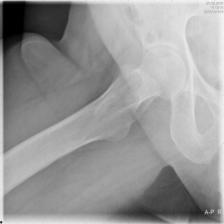}} & \parbox[c]{\linewidth}{\includegraphics[width=\linewidth]{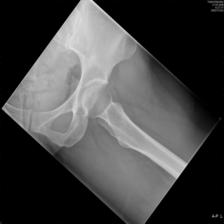}}&\parbox[c]{\linewidth}{\includegraphics[width=\linewidth]{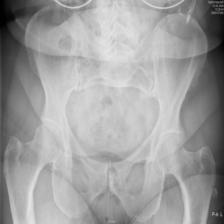}}\\
         \rotatebox[origin=c]{90}{\bf ANKLE} & \parbox[c]{\linewidth}{\includegraphics[width=\linewidth]{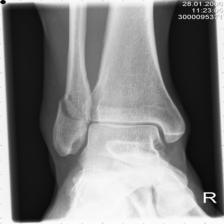}} & \parbox[c]{\linewidth}{\includegraphics[width=\linewidth]{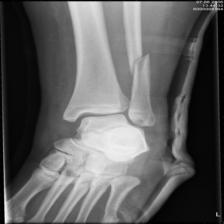}} & \parbox[c]{\linewidth}{\includegraphics[width=\linewidth]{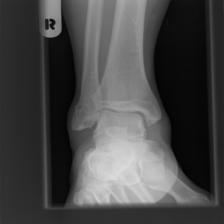}}&\parbox[c]{\linewidth}{\includegraphics[width=\linewidth]{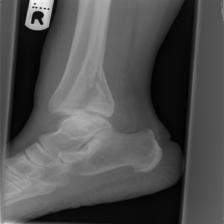}} &  & \parbox[c]{\linewidth}{\includegraphics[width=\linewidth]{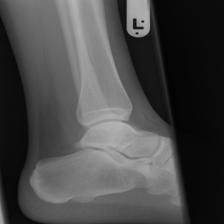}} & \parbox[c]{\linewidth}{\includegraphics[width=\linewidth]{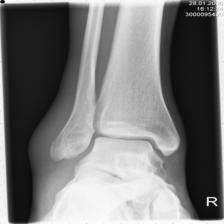}} & \parbox[c]{\linewidth}{\includegraphics[width=\linewidth]{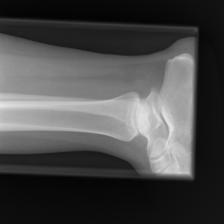}}&\parbox[c]{\linewidth}{\includegraphics[width=\linewidth]{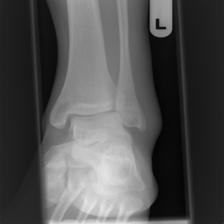}}\\
    \end{tabular}   
    \caption{Example images from medical datasets (resized to $224 \times 224$).}
\label{fig:data_imgs_examples}
\end{figure*} 

\section{Interpretable and Accurate Image Classification}
We evaluate the applicability and potential of PIP-Net for binary medical image classification by first analysing its classification performance~(Section~\ref{ssec_medpipnet_performance_sparsity}), followed by evaluating the interpretability of the learned prototypes~(Section~\ref{ssec:medpipnet_perceived_purity}) and exploring its capabilities to reveal data quality issues~(Section~\ref{ssec:medpipnet_dat_quality}). In Section~\ref{sec:medpipnet_alignment_med} we then evaluate to what extent the learned reasoning is aligned with medical domain knowledge, and the possibilities for manually correcting shortcut learning are investigated in Section~\ref{sec:medpipnet_reveal_correct_shortcut_learning}.

\subsection{Classification Performance and Sparsity}
\label{ssec_medpipnet_performance_sparsity}
\begin{figure}[ht]
\centering
    \includegraphics[width=0.85\linewidth]{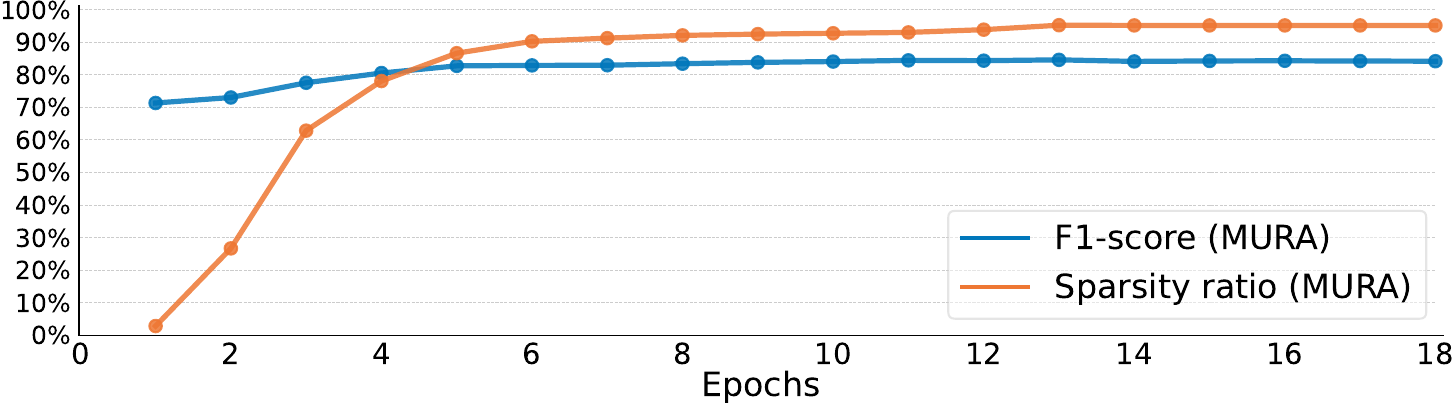}
    \caption{\textbf{MURA} Plot of PIP-Net's sparsity ratio and F1-score during training.}
    \label{fig:med_pipnet_plot_sparsity}
\end{figure}

\begin{table*}[!t]
    \centering
    \caption{PIP-Net's performance on the test sets. Reporting accuracy (Acc), F1 score, sensitivity (Sens), and specificity (Spec). Sensitivity corresponds to the absent-class. Sparsity is the ratio of weights in the classification layer with a value of zero. \# Proto indicates how many prototypes have at least one non-zero weight to a class, and therefore indicates the global explanation size of PIP-Net. The local explanation size indicates the average number of found ($\bm{p}> 0.1$) and relevant (to any of the classes) prototypes in a single test image.}
    \label{tab:med_pipnet_performance}
    \setlength\tabcolsep{4.5pt}%
    \small{
    \begin{tabular}{lccccccc}
    \toprule
        \textbf{Dataset} & \textbf{Acc} & \textbf{F1} & \textbf{Sens} & \textbf{Spec} & \textbf{Sparsity} & \textbf{\# Proto} & \textbf{Local size} \\
        \midrule
        ISIC & 94.1\%& 72.6\% & 97.7\% & 67.0\% & 92.3\% & 119 & 13.8\\
        MURA & 82.1\% & 84.2\% & 91.3\% & 72.0\% & 95.1\% & 75 & 7.6\\
        HIP & 94.0\% & 94.4\% & 93.2\% & 94.9\% & 93.6\% & 99 & 7.4\\
        ANKLE & 77.3\% & 74.0\% & 77.8\% & 76.9\%  & 98.1\% & 29 & 2.5 \\
        \bottomrule
    \end{tabular}}
\end{table*}
Table~\ref{tab:med_pipnet_performance} reports the predictive performance of PIP-Net, and shows that PIP-Net can successfully classify images from all four medical datasets. The ISIC predictive performance is comparable to a standard black box classifier which achieves a sensitivity of 0.90 and a specificity of 0.75~(results from~\cite{nauta_2022_skinshortcut}). Additionally, PIP-Net achieves a high sparsity ratio, thereby lowering the explanation size. These results show that the training mechanism of PIP-Net, originally developed for multi-class classification of natural images~\cite{nauta_pipnet}, also works well for medical binary classification. Fig.~\ref{fig:med_pipnet_plot_sparsity} shows that the sparsity of PIP-Net's classification layer slowly increases during training, while predictive performance is relatively stable. 
Therefore, the number of training iterations mainly influences the trade-off between explanation size and prototype purity (see also next section~\ref{ssec:medpipnet_perceived_purity}).
While a smaller explanation size is generally favourable, it limits the amount of distinct visual concepts that can be represented by separate (pure) prototypes.
The tradeoff between explanation size and prototype purity can be tuned based on visual inspection of the prototypes or automated purity evaluation with part annotations~(as done in the original PIP-Net paper~\cite{nauta_pipnet}).  

\subsection{Perceived Prototype Purity}
\label{ssec:medpipnet_perceived_purity}
\begin{figure*}[!t]
\begin{tabular}{cccc}
    & {\it Fracture / Abnormal}
    & {\it No Fracture / Normal}
    & Pretrained prototypes \\
\rotatebox{90}{\bf ANKLE}
    &\includegraphics[width=0.315\linewidth]{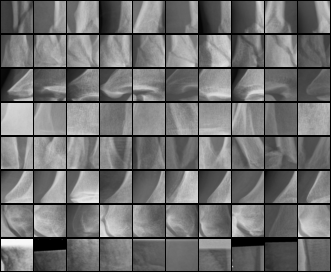}
    &\includegraphics[width=0.315\linewidth]{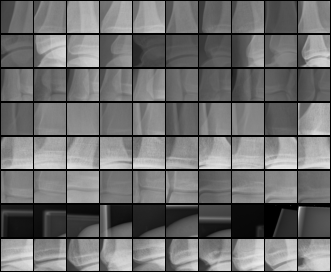} 
    &\includegraphics[width=0.315\linewidth]{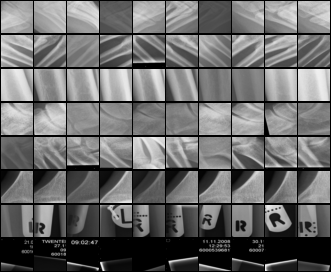}
\\
\rotatebox{90}{\bf HIP} 
    &\includegraphics[width=0.315\linewidth]{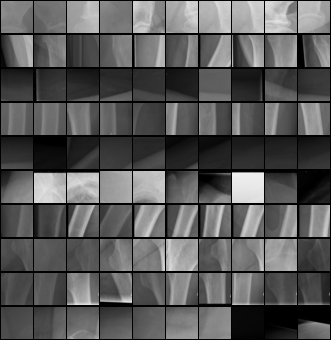}
    &\includegraphics[width=0.315\linewidth]{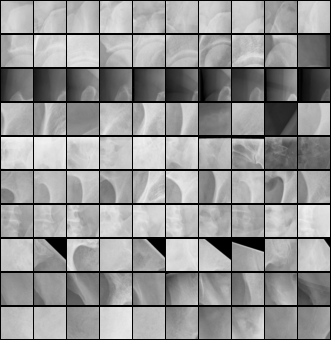} 
    &\includegraphics[width=0.315\linewidth]{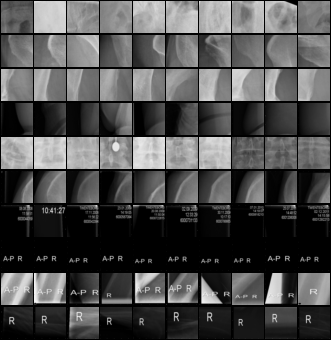}
    \\
\rotatebox{90}{\bf MURA} 
    &\includegraphics[width=0.315\linewidth]{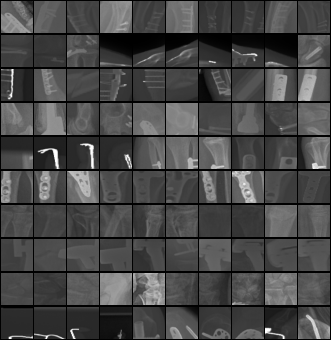}
    &\includegraphics[width=0.315\linewidth]{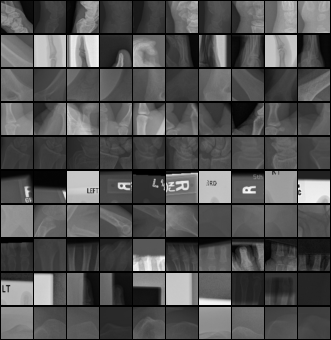} 
    &\includegraphics[width=0.315\linewidth]{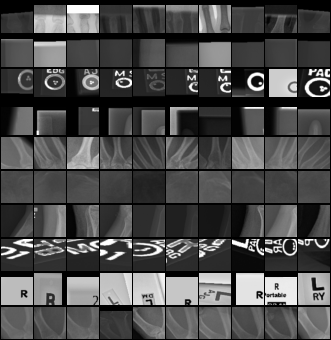}
    \\
\end{tabular}
\caption{Top-10 image patches for the most relevant prototypes per class. One row corresponds to one prototype. \textbf{ANKLE} showing 8 relevant prototypes, since only 8 are relevant for  the \textit{fracture} class. \textbf{HIP} and \textbf{MURA} showing top-10 prototypes. Last column: showing prototypes after pretraining that are eventually not relevant for the classification task, including markers, tags and background. 
}
\label{fig:protype-overview}
\end{figure*}

An important aspect of interpretability of learned medical prototypes, is the purity of the set of image patches corresponding to a prototype. 
A pure prototype clearly represents a single visual concept, such as a metal screw and does not mix multiple concepts, e.g., fractures and metal parts.
We visually assess the prototypes of the MURA, ANKLE and HIP datasets for fractures and abnormalities.
Fig.~\ref{fig:protype-overview} (top and center row), visualises image patches for the most relevant (i.e., highest weight in the classification layer) prototypes for the ANKLE and HIP dataset. It can be seen that image patches for one prototype, either fully trained or only pretrained, look similar, and generally correspond to semantically meaningful concepts. Our findings on medical data are therefore in accordance with high prototype purity for natural images as reported for PIP-Net~\cite{nauta_pipnet}. 

Fig.~\ref{fig:protype-overview} (bottom row) visualises the 10 most relevant prototypes per class for the MURA dataset. Images in MURA are labelled as \textit{normal} or \textit{abnormal} without any additional description. Hence, either the presence of a fracture, hardware lesions or other diseases could be the reason for the \textit{abnormal} label. The part-prototypes of PIP-Net can provide fine-grained insight into the model's learned reasoning. Fig.~\ref{fig:protype-overview}, bottom left, shows that most prototypes relevant to the \textit{abnormal} class represent metal parts, such as operative plates, screw fixation and shoulder prostheses. Prototypes for the \textit{normal} class mainly focus on bones and joints, but PIP-Net also reveals spurious correlations, such as rotated text (cf. Fig.~\ref{fig:protype-overview} bottom center, 6th row). This potential is further investigated in Section~\ref{sec:medpipnet_reveal_correct_shortcut_learning}.

\begin{figure*}[!tbh]
\begin{tabular}{ll}
    \includegraphics[width=0.485 \linewidth]{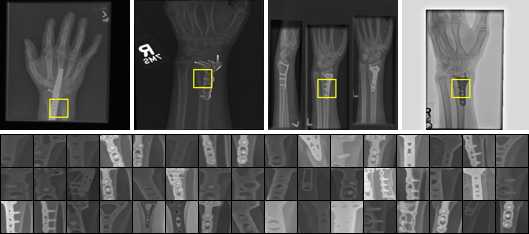}
    &\includegraphics[width=0.485\linewidth]{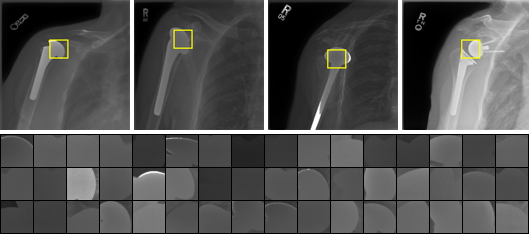}  \\
   (a) Prototype for hand-specific hardware 
    & (b)  Prototype for shoulder prosthesis \\
    \includegraphics[width=0.485\linewidth]{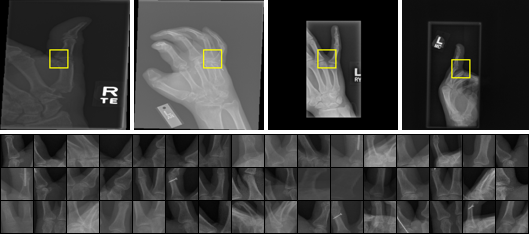}
    & \includegraphics[width=0.485\linewidth]{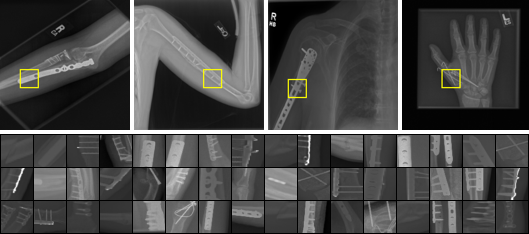} \\
    (c) Prototype for joint disease in finger
    & (d) Prototype for various metalware
\end{tabular}
    \caption{\textbf{MURA} Prototypes relevant to class \textit{abnormal}, visualised with a set of image patches and four representative images indicating where the prototype is detected.}
    \label{fig:mura_combined_all}
\end{figure*}

To facilitate easier interpretation of the prototypes, Fig.~\ref{fig:mura_combined_all} shows representative images from which the prototype image patches are extracted. While the MURA dataset contains images from different body parts, which are shuffled during training, it can be seen that a prototype often relates to a single body part. This is however not always the case, as shown in Fig.~\ref{fig:mura_combined_all}(d) where a prototype represents different types of hardware in different body parts. The optimal trade-off between prototype purity and sparsity can be decided by the user upon visual inspection and tuned with the number of training iterations. 

\subsection{Prototypes for Data Quality Inspection}
\label{ssec:medpipnet_dat_quality}
\begin{figure*}[tbp]
    \centering
\begin{tabular}{p{0.485\linewidth}p{0.002\linewidth}p{0.485\linewidth}}
    \textbf{HIP} & & \textbf{ANKLE} \\
    \includegraphics[width=\linewidth]{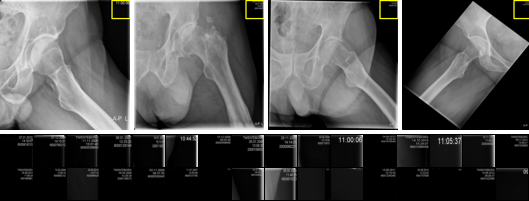}
    & &  \includegraphics[width=\linewidth]{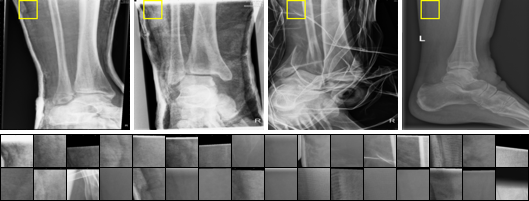} \\
    Pretrained prototype for images with text in the upper-right corner
    & &
    Prototype relevant for the \textit{Fracture} class that represents orthopedic cast and splint
\end{tabular}
    \caption{An inspection of PIP-Net's prototypes can reveal data quality issues.}
    \label{fig:hip_ankle_combined_data_quality}
\end{figure*}

Since PIP-Net first learns prototypes in a self-supervised fashion, 
the pretrained prototypes can inform the user about artefacts and patterns hidden in the training data. Even when these are not discriminative for the prediction task, such as the prototypes shown in Fig.~\ref{fig:protype-overview} (third column), they can facilitate data quality inspection. For example, visual inspection of the prototypes could be used for anomaly detection, or to check whether all individually identifiable information is removed to ensure sufficient anonymisation. Additionally, PIP-Net could assist in quickly identifying labelling errors. For example, the top-left image in Fig.~\ref{fig:mura_combined_all}(a) has the ground-truth label \textit{normal} in MURA but is mislabelled since the presence of metal, a reason for abnormality, is detected. Two examples of data quality issues discovered with PIP-Net are shown in Fig.~\ref{fig:hip_ankle_combined_data_quality}. 
Visual analysis of the pretrained prototypes from the HIP dataset reveals that a few images contain text in the upper-right corner. Similarly, on the ANKLE dataset we find a prototype that relates to orthopedic cast and splint (although it is also activated by a few images with soft tissue, as shown in the rightmost image). X-rays with cast and splints should have been excluded from the dataset as they are taken as part of a follow-up study. Such data collection problems can be easily solved with PIP-Net by removing the images where the prototype is found, in order to prevent that casts and splints become a shortcut for fracture recognition.

\section{Alignment of Prototypes with Domain Knowledge}
\label{sec:medpipnet_alignment_med}
This section evaluates how well the prototypes align with medical domain knowledge. We evaluate prototypes from the ANKLE and HIP datasets by comparing them with medical literature and classification standards, supported by the expert knowledge of a trauma surgeon from hospital ZGT. 
We find that a learned prototype of PIP-Net is generally consistently found at the same location in the body and that most, but not all, of these locations are medically relevant. 

\subsection{ANKLE Dataset}
\begin{figure*}[tb]
    \centering
\begin{tabular}{p{0.485\linewidth}p{0.002\linewidth}p{0.485\linewidth}}
    \includegraphics[trim = {0, 3cm, 0, 10cm},clip,width=0.7\linewidth]{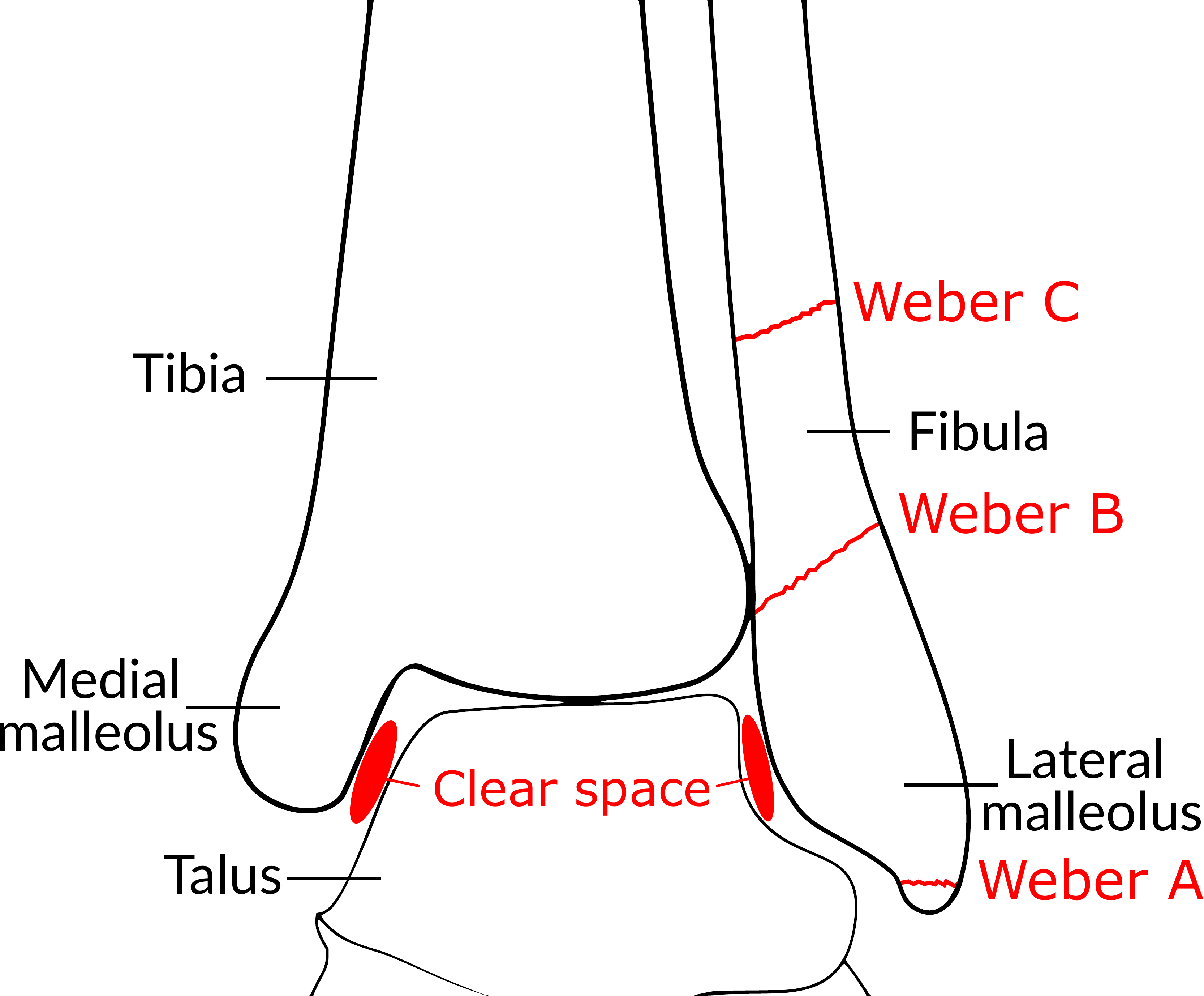} 
    & &  \includegraphics[width=\linewidth]{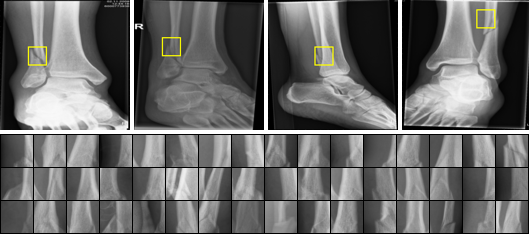} \\
    \includegraphics[width=\linewidth]{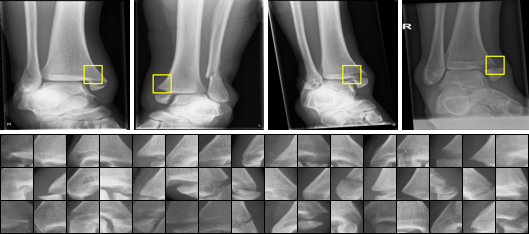}
    & &  \includegraphics[width=\linewidth]{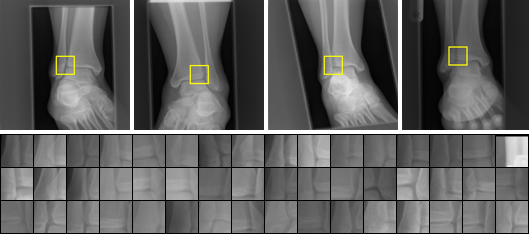} \\
    \end{tabular}
    \caption{\textbf{ANKLE} Correspondence to ankle fracture classification standard, related to three highly relevant prototypes. Visualised with a set of image patches and four representative images indicating where the prototype is detected.\\
    Top left: Weber classification standard; Top right: Prototype relevant to \emph{fracture}, corresponding to Weber B and, with lower presence scores, to Weber C;
    Bottom left: Prototype relevant to \emph{fracture}, consistently found at the distal end of the medial malleolus, a common location for fractures; Bottom right: Prototype relevant to \emph{no fracture}, located at the joint where the fibula, tibula and talus meet.
    }
    \label{fig:ankle_combined_all}
\end{figure*}

Fig.~\ref{fig:ankle_combined_all} visualises a representative subset of PIP-Net's most relevant prototypes for ankle fracture recognition. 
The sketch on the top left illustrates the anatomy of the ankle and additionally indicates the three types of fibula fractures according to the Weber classification standard~\cite{yufit_2010_weber}. 
The most frequently occurring prototype is shown in Fig.~\ref{fig:ankle_combined_all} (top right) and consistently locates fibula fractures. The prototype corresponds to Weber B fractures, which is the most common fracture type~\cite{han2020radiographic}.
The prototype is also activated by Weber C fractures (most-right image), although with a lower prototype presence score. 
The prototype in Fig.~\ref{fig:ankle_combined_all} (bottom left) is consistently found at the distal end of the medial malleolus, which is reasonable as this is a common location for fractures~\cite{han2020radiographic}. These different prototypes show that PIP-Net's reasoning can distinguish between different types of fractures, even though the model is only trained on binary labels. 

Lastly, Fig.~\ref{fig:ankle_combined_all} (bottom right) shows a highly relevant prototype for the \emph{no fracture} class. It focuses on the ankle joint where the tibia, fibula and talus meet, and is only detected in the anteroposterior (AP) views. Another prototype (not shown here) finds the same area in lateral views. It is reasonable that the model checks that this area does not contain any fracture, as Weber B fractures usually end at this joint. Additionally, clear space widening is a radiographic sign which has been shown to be relevant to the diagnosis of ankle fractures~\cite{lau2022understanding}. 
We conclude that PIP-Net learns prototypes for ankle fracture detection that are in line with existing domain knowledge. 
However, we also find that a few prototypes with a lower but non-zero weight are focusing on regions that do not seem to have any medical relevance, such as soft tissue at the top of the foot or ankle. Further future investigation of the prototypes could analyse what discriminative information these image patches hold. 

\subsection{HIP Dataset}
\begin{figure*}[tb]
    \centering
\begin{tabular}{p{0.485\linewidth}p{0.002\linewidth}p{0.485\linewidth}}
     \includegraphics[width=\linewidth]{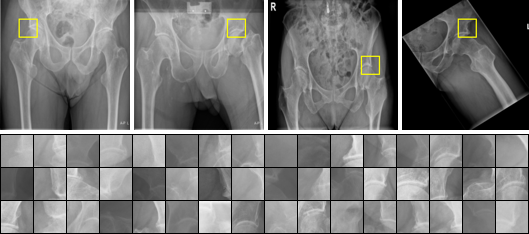}
     & &  \includegraphics[width=\linewidth]{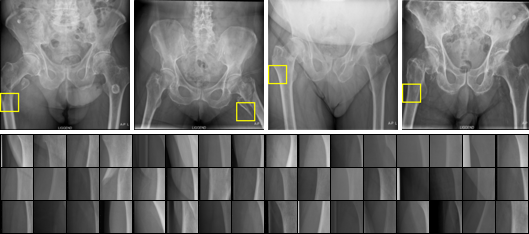} \\
     Prototype consistently found at the joint where the femoral head and the socket meet 
     & & Prototype consistently found at the trochanter minor, an indication for a pertrochanteric fracture\\
\end{tabular}
    \caption{\textbf{HIP} Visualisations of the two most relevant prototypes for the \textit{fracture} class. Visualised with a set of image patches and four representative images indicating where the prototype is detected.}
    \label{fig:hip_combined_all}
\end{figure*}

Fig.~\ref{fig:hip_combined_all} shows the two most relevant prototypes for the \textit{fracture} class. 
The prototype visualised on the left is consistently located at the joint where the femoral head and the socket meet. Although all images contain fractures, the actual fracture is often located slightly below the femoral head. Instead of, or potentially, \emph{in addition to} locating the fracture, the white lines in the image patches might be a sign of arthritis and degeneration of the bone. This indicates a decrease in bone mass and correlates with age, making it a plausible indicator for fractures. The second most relevant prototype, visualised in Fig.~\ref{fig:hip_combined_all} (right), is also consistently found at the same location and is identified in images that include a fracture. The prototype correctly locates the fracture in some images (e.g. two left images showing a pertrochanteric fracture), while focusing below the actual fracture for others (two right images showing a femoral neck fracture). As future work, further investigation can be conducted to identify whether evidence for the fracture is present at these exact locations, or that a learned shortcut is discovered.

\section{Revealing and Correcting Shortcut Learning}
\label{sec:medpipnet_reveal_correct_shortcut_learning}
In this section, we investigate shortcut learning by PIP-Net for the ISIC dataset, which is known to contain coloured patches which spuriously correlate with benign lesions~\cite{icml/RiegerSMY20,nauta_2022_skinshortcut}. Additionally, we investigate a shortcut that was found in the HIP dataset. We evaluate whether the shortcuts can be suppressed by disabling the corresponding prototypes.

\subsection{ISIC Dataset}
\begin{figure*}[htbp]
    \centering
\begin{tabular}{p{0.326\linewidth}p{0.326\linewidth}p{0.326\linewidth}}
    \includegraphics[width=\linewidth]{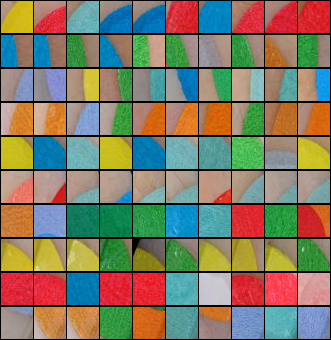}
    & \includegraphics[width=\linewidth]{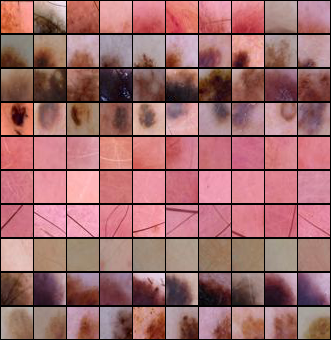}
    & \includegraphics[width=\linewidth]{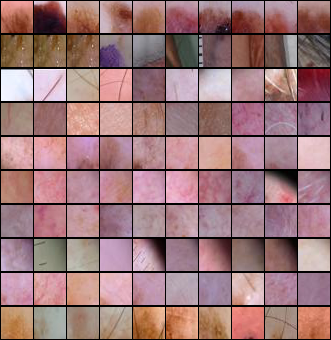} \\
    Relevant to \textit{Benign}, located on coloured patches
    & Relevant to \textit{Benign}, not on coloured patches
    & Relevant to \textit{Malignant} \\
\end{tabular}
    \caption{\textbf{ISIC} Top-10 image patches for the 10 most relevant prototypes per class. One row corresponds to one prototype.}
    \label{fig:isic_prototypes}
\end{figure*}

To evaluate whether PIP-Net bases its decision making process on the presence of coloured patches, a known bias, we first analyse how many learned prototypes correspond to a coloured patch. 
We use the segmentation masks from~\cite{icml/RiegerSMY20} to locate coloured patches and calculate their overlap with image segments where a prototype is detected\footnote{If segmentation masks were not available, patch-related prototypes could efficiently be collected manually, since the sparsity of PIP-Net results in a reasonable number of relevant prototypes (only 119 for ISIC).}.
We label a prototype as related to a coloured patch when at least 20\% of the image patches where the prototype is found (prototype presence score $\bm{p} > 0.1$) has overlap with the patch segmentation mask. We then find that 43 prototypes are related to the spurious coloured patches. Fig.~\ref{fig:isic_prototypes} (left) visualises 10 of these prototypes, which have a positive weight to the \textit{benign} class. 

\begin{table*}[th]
    \centering
    \caption{\textbf{ISIC} Classification performance for different data subsets. Reporting Accuracy (Acc), Sensitivity (Sens), and Specificity (Spec). All images with patches are correctly classified as benign. Inserting spurious patches into malignant images tricks the model into classifying malignant images as benign, revealing unreliable behaviour when used in clinical practice. Disabling patch-related prototypes repairs this behaviour.}
    \label{tab:isic_results}
    \setlength\tabcolsep{2pt}%
    \small{
    \begin{tabular}{l@{\hskip 5mm}ccc@{\hskip 5mm}ccc}
    \toprule
        \textbf{ISIC}&\multicolumn{3}{c}{\textbf{Original PIP-Net}} & \multicolumn{3}{c}{\textbf{Adapted PIP-Net}}\\
         Data subset & Acc & Sens & Spec & Acc & Sens & Spec\\
         \midrule
        Full test set & 94.1\% & 97.7\% & 67.0\% & 93.5\% & 96.7\% & 69.3\%\\
        Excl. images w/ patches & 89.9\%& 95.6\% & 67.0\% & 88.9\% & 93.8\% & 69.3\%\\
        Benign w/ patches & 100.0\% & 100.0\% & n.a.& 99.9\% & 99.9\%& n.a.\\
        Malignant w/ inserted patches & 9.1\% & n.a. & 9.1\%& 65.7\% & n.a. & 65.7\%\\
        \bottomrule
    \end{tabular}}
\end{table*} 
To validate that PIP-Net  bases its decision on these patch-prototypes, we apply PIP-Net to an artificial test set where coloured patches are pasted into \emph{malignant} images, using the same dataset as~\cite{nauta_2022_skinshortcut}. Table~\ref{tab:isic_results} shows that the accuracy for these malignant images (specificity) drops from 67\% to only 9\% when the coloured patch is inserted, confirming that the patch shortcut is indeed exploited. When used in clinical practice, a patient with a malignant lesion could then be misdiagnosed purely because of the presence of a coloured patch. Concluding, these findings motivate the usage of interpretable models: only judging a model based on its predictive performance would not have revealed the shortcut learning. 

\begin{figure*}[th]
    \includegraphics[width=0.47\linewidth]{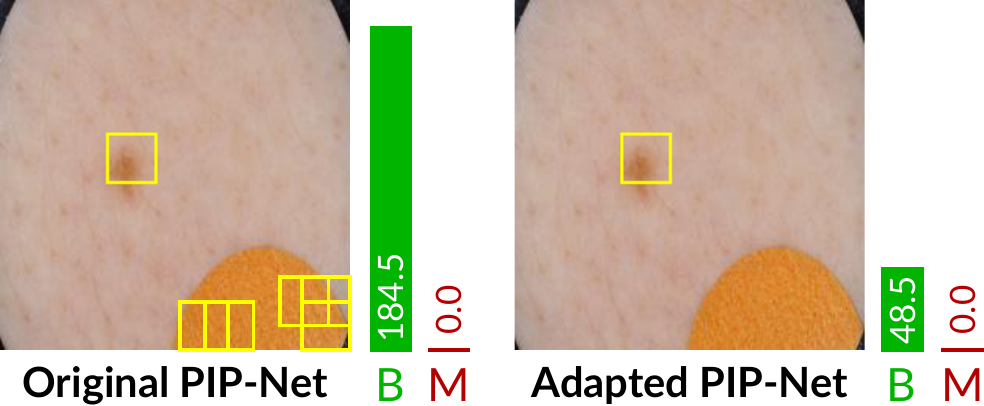} \quad\,
    \includegraphics[width=0.47\linewidth]{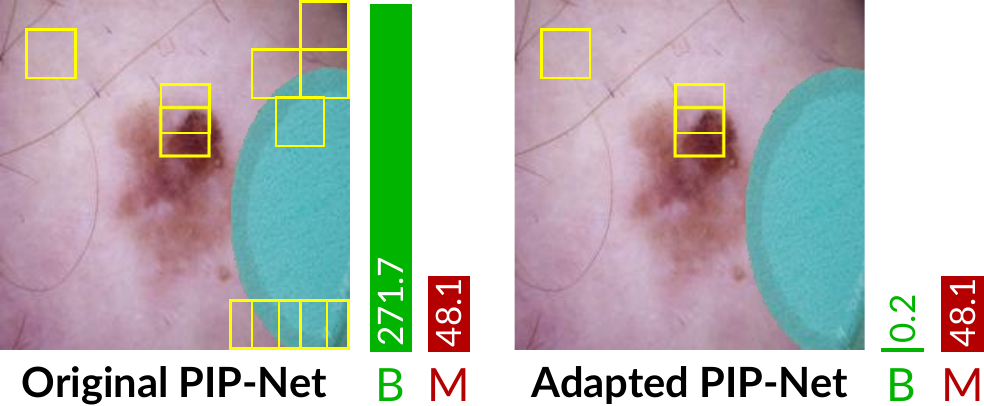} 
    \caption{\textbf{ISIC} Local explanations indicating where prototypes were found, shown for both the original PIP-Net and its adapted version where patch-prototypes are disabled. 
    Left: benign image with patch, right: malignant image with inserted patch.}
    \label{fig:isic_loc_explanations}
\end{figure*}

We investigate whether the shortcut learning in PIP-Net can be corrected by setting the weights of all patch-related prototypes to zero.
Disabling all patch-related prototypes reduces the global explanation size from 113 to 75 prototypes in total. The bottom row of Table~\ref{tab:isic_results} shows that this manual intervention is effective: 
the adapted PIP-Net reaches almost the same accuracy on the adapted malignant patches compared to their original, non-adapted counterparts. 
Additionally, the accuracy of benign images with patches is barely changed. 
Disabling the patch-related prototypes reduces their local explanation size from 13.6 to 2.6 prototypes, but PIP-Net still finds sufficient evidence to classify the benign images correctly. This is supported by the visualised prototypes in Fig.~\ref{fig:isic_prototypes} (centre and right) which indicate that PIP-Net also learned other class-relevant prototypes. In addition to these \emph{global} explanations, Fig.~\ref{fig:isic_loc_explanations} shows two \emph{local} explanations for a test image from both the original and adapted model.

\subsection{HIP Dataset}
\begin{figure*}[th]
 \includegraphics[width=0.485\linewidth]{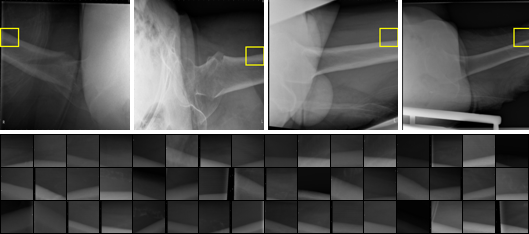}
 \includegraphics[width=0.45\linewidth]{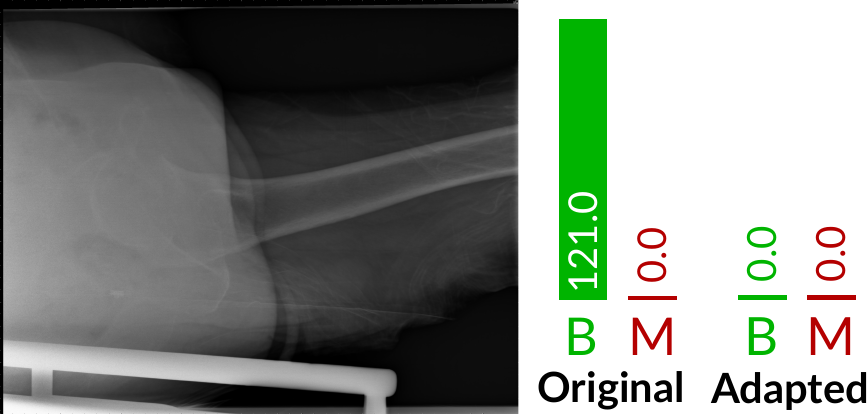}
    \caption{\textbf{HIP} Shortcut learning by PIP-Net for predicting \textit{fracture}. Left: Shortcut prototype that detects a particular view that is only made for immobile patients in the emergency room. In some images, the edge of the bed is still visible. Center: The adapted model does not find evidence anymore for the \textit{fracture} class. Right: output-score for the \textit{fracture} class decreases to zero after suppressing shortcut prototypes.}
    \label{fig:hip_shortcutlearning}
\end{figure*}

Based on visual inspection of the learned prototypes for the HIP dataset, we found that 20 of the 42 prototypes that were relevant to the \textit{fracture} class, were only activated on a specific image view, as shown in Fig.~\ref{fig:hip_shortcutlearning} (left). These views are only taken in the emergency room for immobile patients, and are not part of standard X-ray examinations performed in the outpatient clinic. As immobility is highly correlated with the presence of a hip fracture, it is likely that any deep learning method will use this correlation as a shortcut. We follow a similar approach as for the ISIC dataset, and disable all prototypes that are related to this particular view. Whereas the adapted PIP-Net still found sufficient evidence for correct classifications in the ISIC dataset, it outputs zero scores for roughly half of the images with that specific view, as shown in Fig.~\ref{fig:hip_shortcutlearning} (left). These results indicate that PIP-Net had indeed learned a shortcut, as it was basing its decision for some of these images solely on that particular type of view. Once this shortcut information is suppressed, the output scores will be zero and the model will abstain from a decision. Adapting the model after suppressing shortcut information opens interesting opportunities for future research, such as partial retraining with constraints or integrating human-in-the-loop feedback.

\section{Conclusion}
\label{sec:medpipnet_conclusion}
We demonstrated the wide applicability of PIP-Net by successfully employing it for real-world binary medical image classification.
We see a controlling role for PIP-Net in the medical domain to support human decision making by providing another pair of eyes, underpinned by explanations. Mawatari et al.~\cite{MAWATARI_2020_cnn_radiology} already showed that the diagnostic performance of radiologists for detecting hip fractures in X-ray images improves when they are additionally provided with the prediction of a trained neural network. Rather than only showing predictions, PIP-Net can also show its reasoning based on interpretable prototypical parts. 
Visual analysis showed that image patches corresponding to a learned prototype are coherent and semantically meaningful. We also found that prototypes focus on specific areas that are relevant according to medical classification standards. 

Additionally, PIP-Net can reveal hidden biases and shortcuts in the data that may cause unintended behaviour. The interpretable design of PIP-Net can support the user in quickly identifying these data quality problems. Lastly, we explored whether shortcut learning can be suppressed by directly adapting the reasoning of the model and have shown that disabling prototypes can be an effective way of repairing the reasoning of PIP-Net. 
This capability empowers users not only to uncover improper reasoning, but to take immediate action for correcting it. In contrast, commonly applied black-box models require model adaptations with unclear outcome and costly retraining cycles.

\section{Future Work}
Future research could explore whether retraining the last classification layer \emph{after} disabling the prototypes could improve predictive performance further. We identify this as a promising research opportunity given that the last layer of a standard, uninterpretable neural network can be successfully retrained with only a small subset of the data where the spurious correlation is not present~\cite{kirichenko2022last}. 

Rather than only disabling undesired prototypes \emph{after} fully training the model, we see an interesting research opportunity to have a human \emph{in} the loop who identifies and disables undesired prototypes in an earlier stage, such that the model can adapt itself immediately. It might be even more effective when users can also \emph{add} desired prototypes. When provided with a manual hint, PIP-Net could then automatically refine the prototype in the subsequent learning process. This would allow, for example, to already start with a set of prototypes based on established medical standards. Such an approach would enable a bidirectional feedback loop: an ML model learns prototypes and could start to augment existing medical standards, while medical experts can additionally suggest extra prototypes to the model and disable undesired prototypes. An immediate step towards augmenting existing medical standards could be the detailed inspection of the prototypes found in section~\ref{sec:medpipnet_alignment_med} and whether they indeed contain yet unknown discriminative features. This work is therefore a step towards hybrid AI~\cite{research_agenda_hybrid_akata}, where artificial intelligence complements human intelligence.

\newpage
\bibliographystyle{splncs04}
\bibliography{references}
\end{document}